\documentclass[a4paper,11pt]{article}

\usepackage{fourier}
\usepackage{color}
 \usepackage{graphicx}
\usepackage{url}
\usepackage[affil-it]{authblk}
\usepackage{amsmath}
\usepackage{wrapfig}

\usepackage[T1]{fontenc}
\usepackage{times}
\usepackage{xcolor}

\usepackage{wrapfig,lipsum,booktabs}
\usepackage{graphicx}
\usepackage{subfig}
\usepackage{hyperref}
\usepackage{multirow}
\usepackage[acronym]{glossaries}
\glsdisablehyper

\newacronym{adas}{ADAS}{Advanced Driver Assistance System}
\newacronym{hmi}{HMI}{Human-Machine Interface}
\newacronym{ipm}{IPM}{Inverse Perspective Mapping}
\newacronym{yolo}{YOLO}{You Only Look Once}
\newacronym{bev}{BEV}{Bird-Eye View}
\newacronym{svcs}{SVCS}{Surround-view Camera System}
\newacronym{cvt}{CVT}{Cross-view Transformers}
\newacronym{lss}{LSS}{Lift Splat, Shoot}
\newacronym{mtfcvt}{MT F-CVT}{Multi-Task Fisheye Cross View Transformers}
\newacronym{mlp}{MLP}{Multi-Layer Perceptron}
\newacronym{fov}{FOV}{Field Of View}

\begin{document}

\title{Enhanced Parking Perception by \\ Multi-Task Fisheye Cross-view Transformers}

\author{Antonyo Musabini, Ivan Novikov, Sana Soula, Christel Leonet, Lihao Wang,\\  Rachid Benmokhtar, Fabian Burger, Thomas Boulay, Xavier Perrotton}
\affil{Valeo, Brain Division, Créteil 94000, France}

\date{}
\maketitle
\thispagestyle{empty}

\begin{abstract} \label{section:abstract}
Current parking area perception algorithms primarily focus on detecting vacant slots within a limited range, relying on error-prone homographic projection for both labeling and inference. However, recent advancements in \gls{adas} require interaction with end-users through comprehensive and intelligent \glspl{hmi}. These interfaces should present a complete perception of the parking area going from distinguishing vacant slots' entry lines to the orientation of other parked vehicles. This paper introduces \gls{mtfcvt}, which leverages features from a four-camera fisheye \gls{svcs} with multi-head attentions to create a detailed \gls{bev} grid feature map. Features are processed by both a \textit{segmentation} decoder and a \textit{Polygon--Yolo} based object detection decoder for parking slots and vehicles. Trained on data labeled using LiDAR, \gls{mtfcvt} positions objects within a 25m × 25m real open--road scenes with an average error of only 20 cm. Our larger model achieves an F-1 score of 0.89. Moreover the smaller model operates at 16 fps on an Nvidia Jetson Orin embedded board, with similar detection results to the larger one. \gls{mtfcvt} demonstrates robust generalization capability across different vehicles and camera rig configurations. A demo video from an unseen vehicle and camera rig is available at: \href{https://streamable.com/jjw54x}{https://streamable.com/jjw54x}.

\end{abstract}

\section{Introduction}

\let\thefootnote\relax\footnotetext{This paper is a preprint of a paper submitted to the 26th Irish Machine Vision and Image Processing Conference (IMVIP 2024). If accepted, the copy of record will be available at IET Digital Library.}

\label{section:introduction}
\begin{wrapfigure}{r}{0.5\textwidth}
  \vspace{-60pt}
  \begin{center}
    \includegraphics[width=0.5\textwidth]{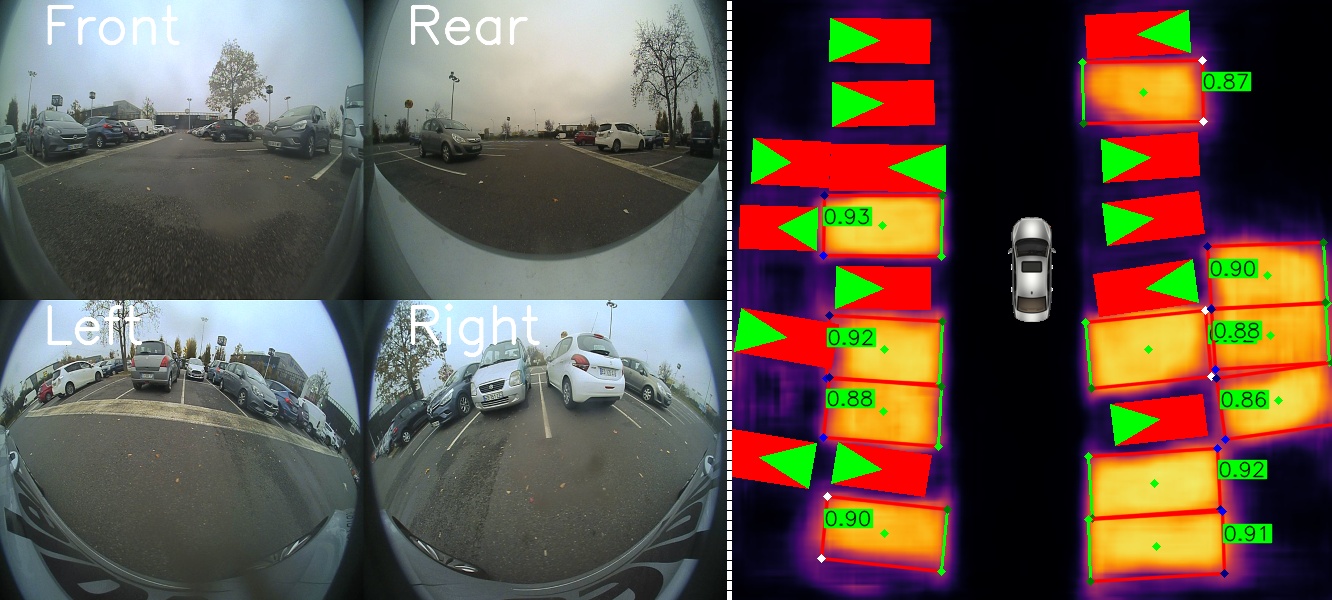} 
    \end{center}
    \vspace{-15pt}
  \caption{\textbf{Left}: Four fisheye images from the surround view camera system. \textbf{Right}: Vacant parking slots and vehicles.}
  \vspace{-10pt}
\end{wrapfigure}

Traditional parking \gls{adas}, found on commercial vehicles, typically address two primary challenges: 1)~detecting and precisely localizing vacant and occupied parking spaces, and 2) autonomously maneuvering into the identified parking slot. Alongside their operational efficiency, systems tackling each challenge require smooth communication with end-users. In response to this requirement, automotive manufacturers have integrated comprehensive \glspl{hmi}, which show parking spots, both vacant and occupied.
The perception algorithm for such \glspl{hmi} should be capable of detecting vacant parking slots holistically and parked vehicles with their correct orientation. Detecting a parking slot holistically involves locating all four corners of the slot and distinguishing its entry line (from where a vehicle enters to that slot), even if a slot is partially visible. Concurrently, there is a perceptual difference between occupied slots and parked vehicles. A vehicle may be poorly parked, being ill-centered within a slot or covering multiple parking slots. Both large and small vehicles can occupy the same slot. Vehicles can be parked forwards or backwards. Hence, the perception algorithm should accurately determine the location, dimensions and the orientation of each detection, even when the ego-vehicle is moving.

Current parking slot detection algorithms often rely on homographic views obtained through \gls{ipm} (i.e. \cite{Zhang2018, Li2020_VPS, Do2020, wang2023holistic}), which introduces errors in size and location of far objects. Therefore, to minimize these errors, the detection range of such algorithms is intentionally kept low. Consequently, by design, the ego-vehicle is unable to perceive a parking slot ahead, whether to illustrate the latter to the driver or to plan an automatic maneuver into it. 
Also, current parking \glspl{hmi}, which aim to meet the real needs of the end-user, typically utilize either only ultrasonic sensors or six pinhole cameras mounted around the vehicle. With that said, vehicles are usually already equipped with four fisheye cameras known as the \gls{svcs}. These wide-lens fisheye cameras, which have a \gls{fov} that can go up to 195$^{\circ}$, 
cover the vehicle’s surroundings without leaving any blind spots. Although this large \gls{fov} is practically limited by the ego-vehicle’s body, each camera still covers important overlapping zones from at least two other cameras, making \gls{svcs} suitable for use in a cross-view architecture. Moreover, their low mounting positions make them ideal for parking-related applications. Another advantage worth citing is that with \gls{svcs} we mount less cameras (two fewer) than the already mentioned pinhole set-up, making it more cost-effective, computation-effective and data effective. However, due to the lens distortion inherent in fisheye cameras and the lack of open datasets with fisheye images, their use in perception systems is often overlooked compared to the six-pinhole camera configuration.

The research question we aim to answer in this work is as follows: \textit{Can a more advanced \gls{bev} space feature projection, taking into account the fisheye camera-geometry, using all the cross-views together with an attention mechanism, resolve the presented issues, ensuring real-time execution speed on embedded hardware?} Therefore, this work presents a novel parking perception algorithm that utilizes only the \gls{svcs}: \gls{mtfcvt}. The features extracted from four fisheye cameras are projected into a \gls{bev} feature space using cross-view transformers. Additionally, the algorithm is trained with real 3D annotations obtained from a LiDAR, which eliminates errors induced by perspective projection. By applying cross-view transformers to create the \gls{bev} feature space, and integrating real 3D annotations, we aim to leverage the data fusion from all cameras to overcome the limitations of homographic view-based approaches. Furthermore, atop the created \gls{bev} feature space, segmentation and object detection tasks are designed in a multi-task approach. 
\gls{mtfcvt}, accurately locates holistic vacant parking slots with an additional flag per corner indicating whether they are directly perceived from the cameras or inferred theoretically and identifies vehicles along with their orientation.

\section{Related Work} \label{section:sota}




\paragraph{Parking Slot Detection.}  \label{section:sota:parking_slot_detection}


Recent computer vision and deep learning-based solutions have managed to address primarily detecting vacant slots using convolutional networks~\cite{Zhang2018}
or with attentional graph networks~\cite{min2021attentional}. 
Although they are capable of identifying empty parking spaces, they are all limited by their reliance on the homographic view. This view is used to generate a top-view image, which is the input to the network both during training and inference. Due to the inherent assumption of ground flatness in such approaches, these networks have difficulty accurately positioning the detected slots and their dimensions may vary depending on the distance from the ego-vehicle. Additionally, the use of the homographic view requires stitching between camera views, which introduces non-continuous ground markings on the RGB top-view and varying lighting conditions. These are the main reasons why previous work kept the parking detection range intentionally low ($\pm5m$ around the vehicle). \cite{wang2023holistic} introduced the concept of polygon shapes for parking slots, which is particularly beneficial in conditions of low visibility. With this approach, they extended the detection range to $\pm12.5m$ around the vehicle but did not address the issues associated to the homographic view. As a matter of fact, due to the homographic view, high objects become visually enlarged, making it difficult to detect objects other than vacant slots (i.e. parked vehicles and their orientation, pedestrians). 

\paragraph{Bird-Eye View Projection Methods.} \label{section:sota:bev}
Recently, various \gls{bev} space feature projection methods have been introduced in the following categories: 1) homography-based, 2) depth or geometry-based, 3) MLP-based, and 4) attention-based approaches. Homography-based approaches for parking detection are discussed in Section~\ref{section:sota:parking_slot_detection}. The pioneer depth-based projection, \gls{lss}~\cite{philion2020lift}, first creates a latent depth distribution for image features. Then, the frustums of features from each independent camera are rasterized to a \gls{bev} space feature grid. 
With a similar projection method, M$^{2}$BEV \cite{xie2022m} suggests the multi-task learning of a segmentation and an object detector, and concludes that joint training slightly degrades individual performances of each task. We argue that using a cross-view attention mechanism can overcome this unexpected result as the learnt feature projection is not based on a single camera view. 
MLP-based methods are known to be camera-dependent as they neglect camera intrinsic and extrinsic parameters \cite{ma2022vision}. While being effective, attention-based approaches are in general known to be computationally costly (i.e.\cite{bartoccioni2022lara}). 

\gls{cvt}~\cite{zhou2022cvt} use the multi-head cross-attention mechanism for learning a costly dense mapping between multi-camera features and the BEV grid pixels. \gls{cvt} uses down-sampled features from image feature extractors, which contributes to it's real-time execution speed (each down-sampled level is called a backbone endpoint). Despite their \gls{bev} feature projection capabilities, all of the presented methods take six pinhole cameras as input but ignore fisheye camera geometry. 

F2BEV~\cite{samani2023f2bev} generates height maps and \gls{bev} semantic segmentation map 
using fisheye camera geometry. However, they used only synthetic images created from a 3D simulator for training and inference.
BEVFastLine~\cite{narasappareddygari220bevfastline} processes real \gls{svcs} images with a \gls{bev} architecture only for detection of parking line detection, disregarding holistic parking slots, occupied slots and parked vehicles. NVAutoNet~\cite{pham2024nvautonet} presents a multi-task network, where one of the task is a parking slot detector. However, they use an 8 camera configuration: 4 \gls{svcs} and 4 additional pinhole cameras. Evaluated on a private dataset, they report identical performances for single-task vs. multi-task training for parking slot detection.


To the best of our knowledge, our architecture is the first work which projects only features from real fisheye camera captures into a \gls{bev} grid for learning simultaneously, multi tasks tailored for a complete parking perception.

\section{Proposed method} \label{section:methodology}


\subsection{Global Architecture} \label{section:methodology:global}





\begin{figure}[!htb]
    \vspace{-10pt}
    \centering
    \includegraphics[width=.8\linewidth]{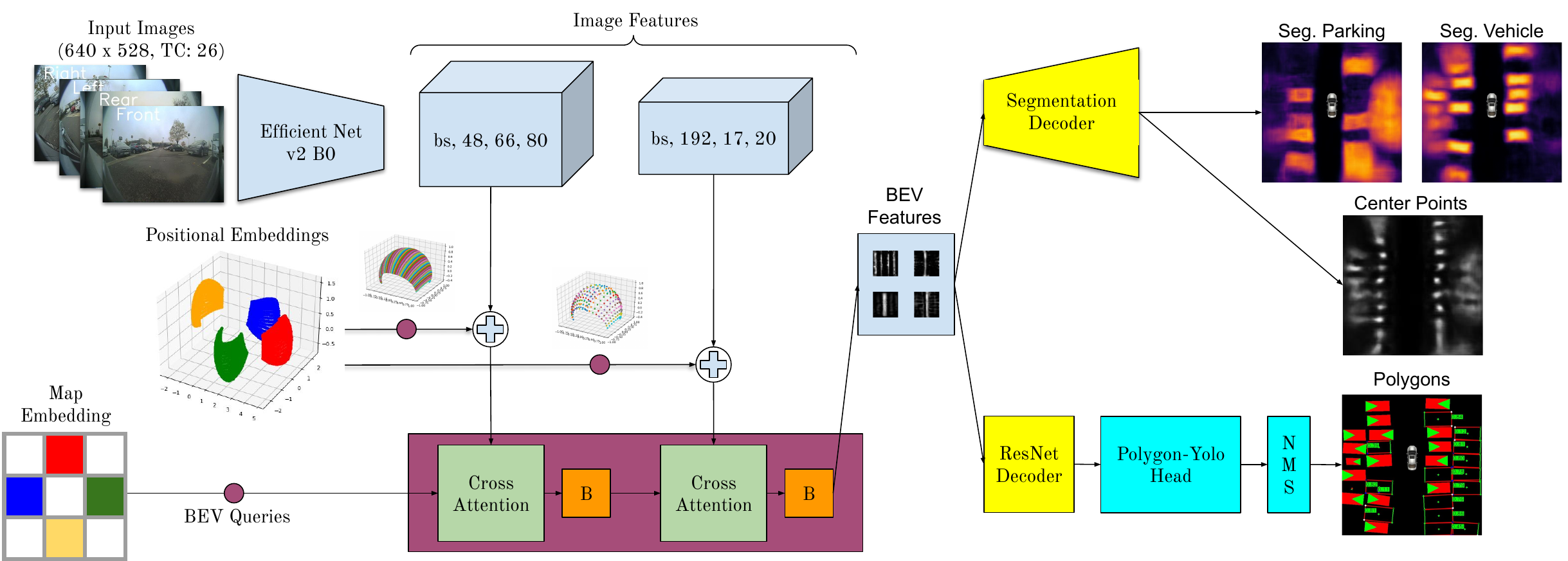}
    \caption{\textbf{Global architecture.} On the left-upper part; input images, the features extractor and its output dimantions are visible. In the left-middle, the fisheye-aware positional embedding and their down-scaled shapes are illustrated. On the left-bottom, position-aware map embedding, the multi-head cross-view transformers and the bottlenecks are shown. On the right side, based on the BEV features, the two independent task heads and their respective outputs are depicted. Purple circles represents \gls{mlp} layers.}
    \label{fig:global_architecture}
    \vspace{-10pt}
\end{figure}


Figure~\ref{fig:global_architecture} presents the global architecture of \gls{mtfcvt}, which is based on~\cite{zhou2022cvt}. This base architecture is chosen for the presented need of using the cross-views all together with an attention mechanism and for its fast execution linked to the down-sampled features (called endpoints). An attention function is described as mapping a query and a set of key-value pairs to an output \cite{vaswani2017attention}. In Figure~\ref{fig:global_architecture}, queries are shaped as a map embedding, key-value pairs are the fisheye positional embedding - image features pairs and the attention outputs are the \gls{bev} features. Originally, the architecture was designed for six-pinhole cameras. It was also equipped with an EfficineNet B4 backbone~\cite{pmlr-v97-tan19a}. The dimensions of the features were down-scaled 4 times (endpoint 2) and 16 times (endpoint 4) before being passed into cross-attention modules with multiple heads. First, we propose replacing the backbone with versions of EfficineNet v2~\cite{pmlr-v139-tan21a}, as it optimizes training speed and parameter efficiency with new operations such as Fused-MBConv.

We present two architectures, one aimed at meeting real-time constraints and the other at achieving the best performance. Input images are resized to 640 × 528 with a top-crop of 26 pixels, as the upper parts of the images contain less relevant information for parking-related scenes. Our smaller architecture uses EfficientNet v2 B0 with feature dimensions down-scaled 8 times (endpoint 3) and 32 times (endpoint 5), while the larger one uses the EfficientNet v2 L backbone with endpoints 2 and 4. 
Early endpoints output feature maps with fewer channels but with higher dimensions, whereas late endpoints leverage the backbone's deeper features, projecting more channels with reduced dimensions. Features are fused with the positional embeddings of the fisheye cameras (see Section~\ref{section:methodology:fisheye}). The results are projected onto a \gls{bev} feature space, with each feature set processed by a 4-head cross-attention of 32 channels (resulting in 128 \gls{bev} feature channels). The \gls{bev} features output a grid size of 25 × 25, covering a total area of 25m × 25m around the vehicle. Subsequently, this common representation is processed by different independent task heads (see Section~\ref{section:methodology:tasks}).

\subsection{Fisheye Positional Embeddings} \label{section:methodology:fisheye}

\begin{wrapfigure}{r}{0.5\textwidth}
  \vspace{-50pt}
    \centering
    \subfloat[\centering Six pinhole]{\includegraphics[trim=75 30 75 100, clip, width=.40\linewidth]{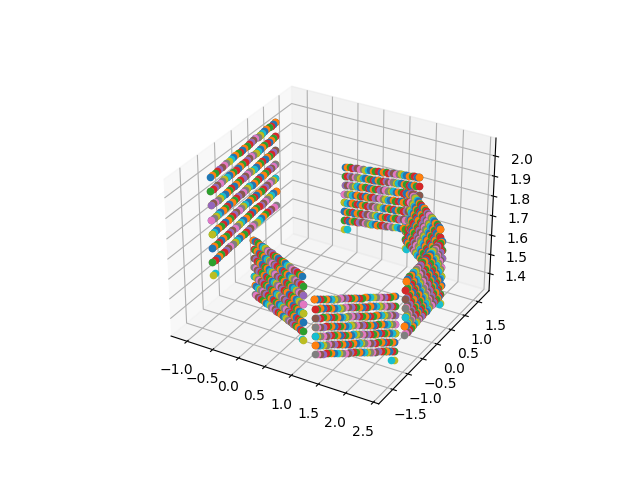}\label{fig:projections:_nuscenes}}
    \qquad
    \qquad
    \subfloat[\centering Four fisheye]{\includegraphics[trim=75 30 75 100, clip, width=.40\linewidth]{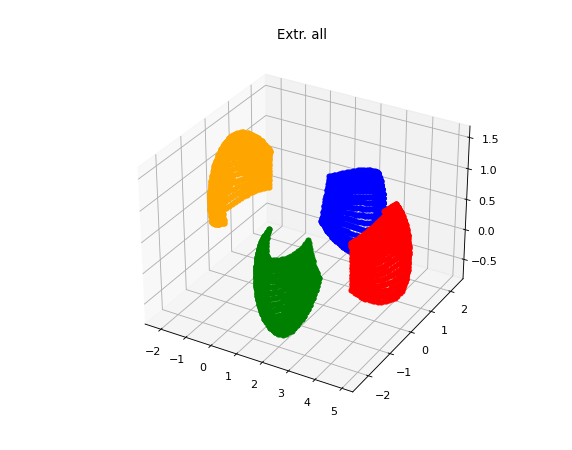}\label{fig:projections:svcs}}
    \qquad
    \vspace{-5pt}
    \caption{\textbf{Projection encoders.} a) Pinhole cameras, as flat surfaces, in respect of their \glspl{fov}. b) Fisheye cameras, in respect to lens's radial distortion ({\color{red}{front}}, {\color{blue}{left}}, {\color{yellow}{rear}}, {\color{green}{right}}).}
    \label{fig:projection_encoders}
    \vspace{-10pt}
\end{wrapfigure}

Following the equidistant fisheye geometry model, the radial distortion of fisheye lens causes an incident angle $\alpha$ linked to it's radial euclidean distance $r_d$, in the distorted image plane, centered in the principal point. In order to mapping pixel positions in image coordinate frame to it's incidence angle, Polynomial FishEye Transform (PEFT) is used by computing the root $\alpha$ of $r_{d,PEFT}=c_1\alpha+c_2\alpha^2+c_3\alpha^3+c_4\alpha^4$. The parameters $c_k$ are obtained during the camera calibration. Figure~\ref{fig:projection_encoders} illustrates the projection encoders, after being multiplied with the inverse of the extrinsic parameters, for both the initial six-pinhole camera configuration from nuScenes dataset~\cite{nuscenes} (Figure~\ref{fig:projections:_nuscenes}) and our adapted four-fisheye camera configuration (Figure~\ref{fig:projections:svcs}). In both figures, the rear axle of the vehicle is located at the origin point $(0,0)$. The pinhole cameras result in a flat plane projection and regular distances between pixels (with the rear camera having a broader \gls{fov}), while the fisheye cameras produce curved projections with variable $\alpha$ angles between rays, due to the radial distortion of the camera lenses. Projection encodings are processed by \gls{mlp} layers before being used as positional embeddings into the down-scaled features. This embedding provides positional information for aggregating features from different cameras, as \textit{attention keys} described earlier.


\subsection{Multi-Task Heads} \label{section:methodology:tasks}

\gls{mtfcvt} executes two separate tasks in parallel, using a shared \gls{bev} grid feature map, followed by distinct decoders. These tasks are implemented through a \textit{segmentation} head and a \textit{Polygon--Yolo} head (refer to Figure~\ref{fig:global_architecture}). The \textit{segmentation} head is based on the original implementation by~\cite{zhou2022cvt}. It decodes and up-samples the \gls{bev} grid three times successively to produce segmentation maps and center point maps for parking slots and vehicles. These outputs are considered auxiliary in our implementation, as they can't identify neither vehicle orientations nor parking slot entry lines. The object decoder head uses three ResNet blocks as individual decoders. Upon these blocks, an enhanced version of the \textit{Polygon--Yolo} head~\cite{wang2023holistic} is used, which predicts the four corners of polygon-shaped objects and their classification confidence. Additionally, we predict a flag for each corner, called corner visibility, indicating whether it is directly perceived by the vehicle or is inferred without direct visibility, as an additional feature.

\section{Experiments} \label{section:Experiments}

\subsection{BEV Dataset for Parking \& Vehicles}  \label{section:methodology:datasets}





\begin{wraptable}{r}{0.5\textwidth}
    \vspace{-70pt}
    \begin{center}
    \begin{tabular}{c|c|c|cccc||c|ccc}
        \hline
        \textbf{Split} &
        \textbf{N. Image} &
        \textbf{Parking Slots} &
        \textbf{Vehicles}
          \\ \hline
        Train & 
        38,028 & 
        334,509 & 115,345 \\ \hline
        Val & 
        10,261 & 
        95,809 & 31,400 \\ \hline
    \end{tabular}
    \end{center}
    \vspace{-15pt}
    \caption{\textbf{Dataset.} Labels within a range of 25m around the vehicle, with a maximum relative occlusion of 70\%.} \label{table:dataset:v_2_9_0}
    \vspace{-10pt}
\end{wraptable}

In order to train our model, we use an private dataset of fisheye images captured in different climate conditions, countries and speeds. Table~\ref{table:dataset:v_2_9_0} details their distribution. Labels were annotated directly in the \gls{bev} space using a LiDAR mounted on the vehicle. Vehicle odometry is used to fill in labels for a given frame when slots are not completely visible. In fact a binary visibility flag is generated per parking slot corner. Thanks to this \gls{bev} annotation method, the ground truth data is not affected by potential errors related to homographic view stitching, camera lens distortions, or inaccurate 3D location estimation. The multi task training of both heads 
was carried out under a two-class problem: \textit{parking} (parallel, perpendicular and angles slots) and \textit{vehicle} (cars, two wheeler, buses, etc.) (see Figure~\ref{fig:qualitative_results}). 

\subsection{Implementation Details} \label{section:implementation_details}

\begin{align}
    \mathcal{L}_{MT-FCVT} = \mathcal{L}_{binary\ seg.\ loss} + 1e^{-1} * \mathcal{L}_{seg.\ center\ loss} + 
    5e^{-2} * (1-\mathcal{L}_{polygon\ GIoU}) + \nonumber\\
    7.5e^{-1} * \mathcal{L}_{bce.\ objectness} + 6.25e^{-3} * \mathcal{L}_{bce.\ class} +   5e^{-2} * \mathcal{L}_{corner\ distance} + 3e^{-3} * \mathcal{L}_{bce.\ corner\ visibility}
    \label{eq:loss}
\end{align}

All training sessions were conducted with a batch size of 8, with AdamW optimizer. 
A one-cycle learning rate policy was applied, with a maximum learning rate of 3e-4 achieved at the midpoint of a total of 400k steps (starting value: 1.5e-4, ending value: 1.5e-5). Eq.~\ref{eq:loss} depicts the weighted distribution of losses between tasks. Segmentation and center losses are binary sigmoid focal losses \cite{lin2017focal}, computed per output channel, where each channel represents a single class (as in \cite{zhou2022cvt}). Objectness and object class losses are binary cross entropy with logistic losses. Polygon Generalized-IoU and corner distance losses follow their implementation from ~\cite{wang2023holistic} and finally our proposition (i.e. the corner visibility loss) is another binary cross entropy with logits loss, used for only parking slots corners. Training of the \gls{mtfcvt} lasted approximately 4 days on an Nvidia A100 GPU. For each backbone, their provided pre-trained weights were used. 


\subsection{Results}

\paragraph{Quantitative Results.}

\begin{table}[!htb]
    \vspace{-5pt} 
    \begin{center}
    \begin{tabular}{c|ccc|cccc|cc}
        \hline
        &
        \textbf{B.bone} &
        \textbf{Endpts.} &
        \textbf{Parameters} &
        \multirow{2}{*}{\textbf{F-1 Sc.}} &
        \multirow{2}{*}{\textbf{Prec.}} &
        \multirow{2}{*}{\textbf{Recall}} &
        \textbf{{Dist. Err.}} &
        \multicolumn{2}{c}{\textbf{Execution Speed (FPS)}}
        \\
        & \textit{Eff.net} & \textit{1st / 2nd} &
        \textit{Millions} & 
        &  & & \textit{(cm)} &
        \textit{1080Ti $^\textit{fp32}$} &
        \textit{J. Orin $^\textit{fp16}$} \\
        \hline
        H & - & - & - & 
        .660 & .59 & .75 & - & - & - \\  \hline \hline
        i & B4 & 2 / 4 & 14.6 & 
        {\color{red}.829} & .79 & .87 & 26 & 8.2 & 7 \\ \hline 
        - & v2 B0 & 2 / 4 & 15.7 & 
        .833 & .79 & .88 & 26 & 19.3 & 10 \\ \hline
        s & v2 B0 & 3 / 5 & 11.3 & 
        .859 & .84 & .88 & 23.1 & \textbf{{\color{cyan}19.6}} & \textbf{{\color{cyan}16}} \\ \hline
        L & v2 L & 2 / 4 & 35.8 & 
        \textbf{{\color{cyan}.884}} & .88 & .89 & 20 & 5.6 & {\color{red}4} \\ \hline        
        - & v2 L & 3 / 5 & 122 & 
        .882 & .88 & .89 & 19.1 & {\color{red}5.1} & {\color{red}4} \\      
        \hline
    \end{tabular}
    \end{center}
    \vspace{-15pt}
    \caption{\textbf{Architectural ablation table.} Comparison of various backbones and used endpoints, in terms of detection metrics and runtime performances. Each model benefit from cumulative contributions of configuration $\star$ from Table~\ref{table:Results_Ablation}. $H$: homograph--stitching based method from \cite{wang2023holistic}, $i$: initial configuration from \cite{zhou2022cvt}, $s$: small configuration, $L$: large configuration. $\textit{fp32}$: direct execution of models including the auxiliary outputs. Detection performances reflect this configuration. $\textit{fp16}$: execution of a simplified ONNX in float 16 precision. Auxiliary outputs are omitted during ONNX execution and Parameters column reflects this configuration.} \label{table:Results_Ablation_Bbone}
    \vspace{-5pt}    
\end{table}


Ideally, comparing \gls{mtfcvt} to state-of the-art methods would be insightful. However due to the nonexistence of any publicly available parking area datasets, annotated in \gls{bev} with the presence of \gls{svcs}, our quantitative evaluation focuses on our internal dataset. Table~\ref{table:Results_Ablation_Bbone} shows free slot and vehicle detection performance to speed (frames per second) for various backbones and endpoints. The confidence threshold is set to 10\%. By the design of Polygon Generalized-IoU loss, objects are considered well detected, when both their position and orientation are acuratly predicted. It is noticeable that, both \textit{small} and \textit{large} proposed models perform better than the \textit{initial} configuration, in terms of F1 score. The distance error reflects the mean positioning error of the two heading points (both for vehicles and parking slots). \textit{Small} achieves positioning objects with a mean error of 23.1 cm.  Also, it achieves 16 fps execution speed on an Nvidia Jetson Orin embedded board. 
It is noticeable for such a multi-head attention mechanisms, using late endpoints of a lighter backbone is beneficial both in terms of detection metrics and execution speed, whereas, for a large backbone, it leads to a slower execution speed without significant improvement in detection performance. Finally, retraining of the homograph--stitching based method from~\cite{wang2023holistic} is proceeded under our two class problem (line H from Table \ref{table:Results_Ablation_Bbone}). As expected, it presents the weakest detection performance among all tested configurations.

\begin{figure*}[!htb]
    \vspace{-10pt}
    \centering
    \subfloat[\centering Perpendicular parking slots.]{\includegraphics[width=.30\linewidth]{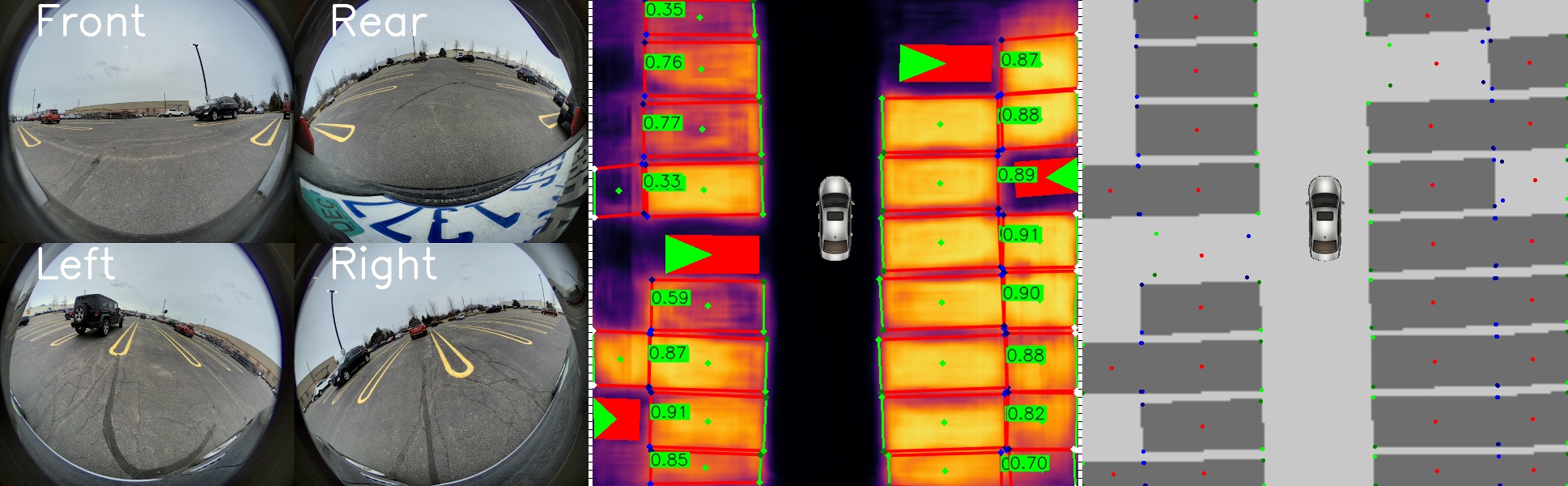}\label{fig:0064}}
    \qquad
    \subfloat[\centering Ego-vehicle in a slot.]{\includegraphics[width=.30\linewidth]{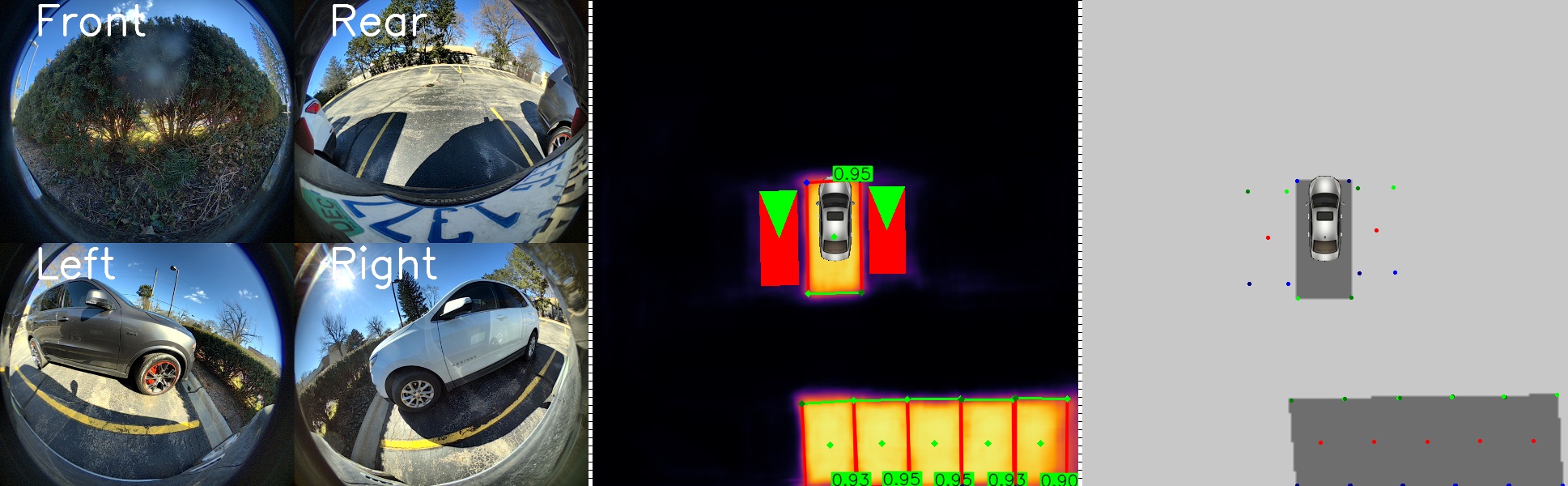}\label{fig:0103}}
    \qquad
    \subfloat[\centering Parallel parking slots]{\includegraphics[width=.30\linewidth]{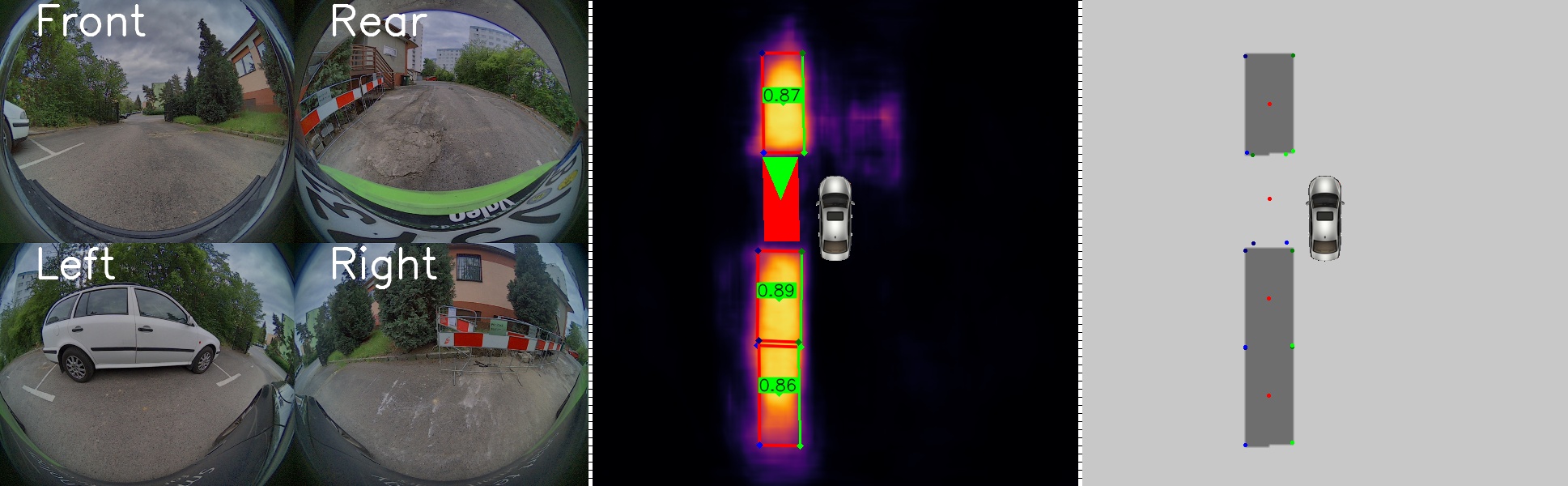}\label{fig:0127}}
    \qquad
    \subfloat[\centering Inferred without direct visibility.]{\includegraphics[width=.3\linewidth]{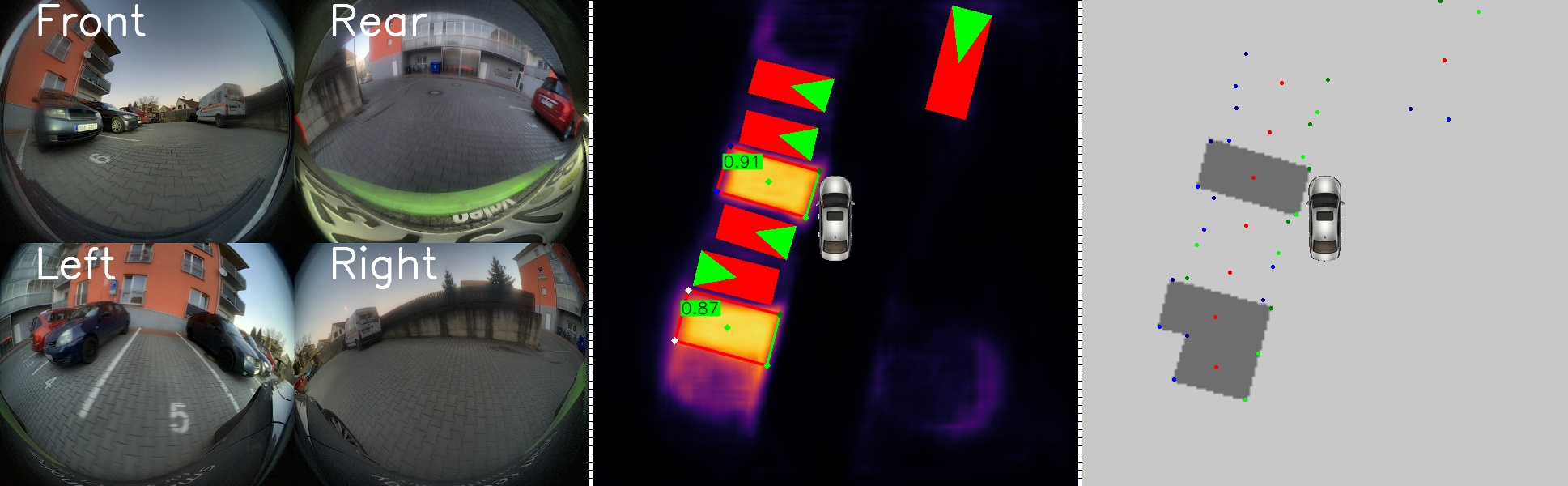}\label{fig:0192}}
    \qquad
    \subfloat[\centering Indoor parking area.]{\includegraphics[width=.3\linewidth]{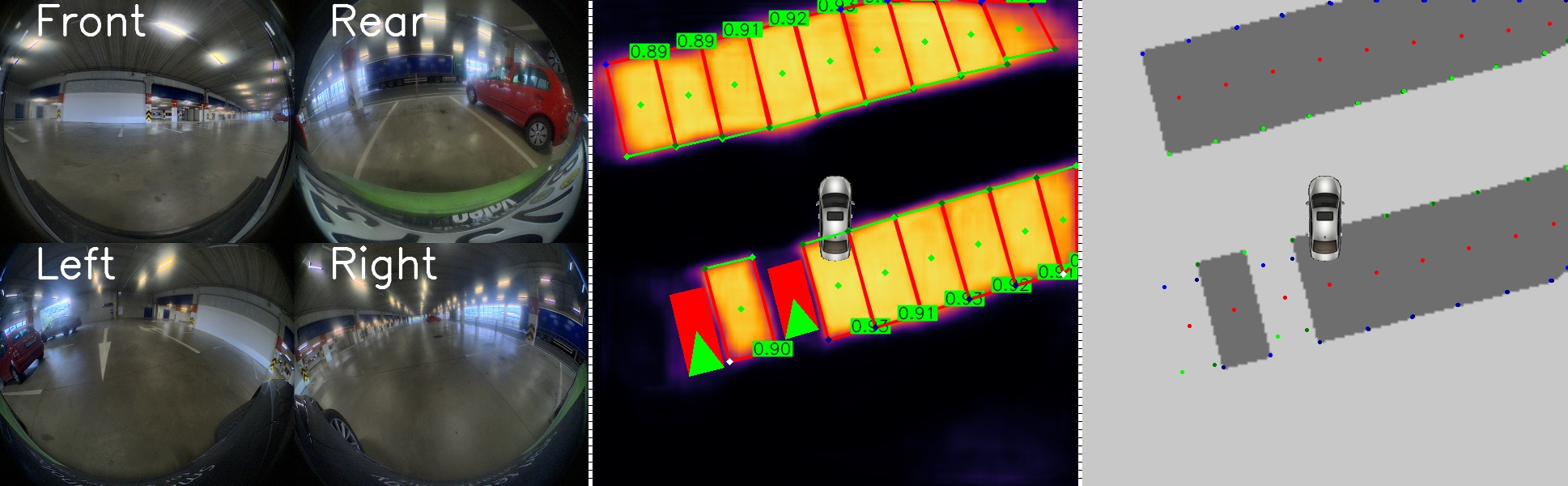}\label{fig:0238}}
    \qquad
    \subfloat[\centering Parallel parked vehicles]{\includegraphics[width=.3\linewidth]{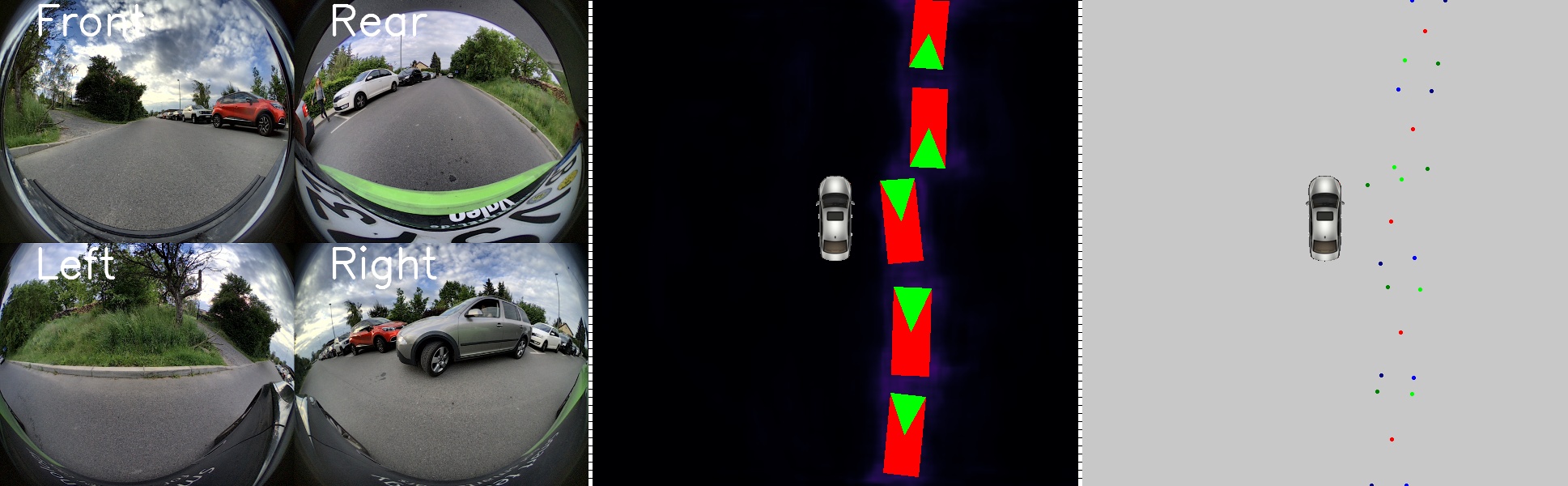}\label{fig:0585}}
    \qquad
    \caption{\textbf{Qualitative results.} Inference examples where the four fisheye images (on left), the predictions (on middle) and the annotated labels (on right) are visible. For labels, the dark gray zones represent the segmentation maps of parking areas (the segmentation map for vehicles is not illustrated for convenience). The polygon labels for both vehicles and parking areas are also visible. The red dots indicate their center positions, the green dots correspond to their heading directions (entry line for parking areas), and the blue dots mark their rear sides.
    }
    \label{fig:qualitative_results}
\end{figure*}

\textit{Large} predicts the corner visibility flag of each slot corner with .93 accuracy. This additional output is used to select the appropriate color for parking corners during inference time (see white points from Figure \ref{fig:0192}).


\begin{wraptable}{r}{0.5\textwidth}
    \vspace{-20pt}
    \begin{center}
    \begin{tabular}{c|c}
        \hline
        \textbf{Description} &
        \textbf{F-1 Sc.}
        \\
        \hline
        $\|$ Single task, Fisheye-LSS projection & .603  \\ \hline \hline 
        $\mathparagraph$ Single task, Fisheye-CVT projection & .849  \\ \hline
        \textbf{+} Image top crop, 26 pixels ($\sim5\%$) & .855  \\ \hline  
        \textbf{+} Multi-Task learning & .864   \\ \hline
        \textbf{+} Image color space and noise & .865 \\ \hline 
        \textbf{+} Roll rotation per camera ($\pm10^{\circ}$, p=.9) & .865 \\ \hline
        $\ddagger\ddagger$ \textbf{+} BEV features droupout, p=.5 & .872  \\  \hline        
        $\ddagger$ \textbf{+} BEV flip left-right (F-LR), p=.5 & .886    \\  \hline
        $\mathsection$ \textbf{+} BEV yaw rotation $\pm22.5^{\circ}$ , p=.9 & \textbf{.891}    \\  \hline \hline

        $\dagger$  (+ from $\ddagger\ddagger$) BEV F-LR, \\
        yaw: $90^{\circ}$-$180^{\circ}$-$270^{\circ}$, p=.2 each &.880    \\  \hline 
        $\star$  + BEV yaw rotation $\pm22.5^{\circ}$ , p=.9 &.884  \\        \hline \hline

    \end{tabular}
    \end{center}
    \vspace{-15pt}
    \caption{\textbf{Methodical ablation table.} Cumulative contributions for \textit{large} model. p: probability. +: on the top of the previous row.} \label{table:Results_Ablation}
    \vspace{-15pt}
\end{wraptable}

Table~\ref{table:Results_Ablation} presents an ablation study of various techniques used during this work. First, in order to study the impact of the fisheye camera feature projection method, we compare the attention based single task fisheye-CVT ($\mathparagraph$) with fisheye adaptation of the pioneer attention-free \gls{bev} projection method: LSS ~\cite{philion2020lift} ($\|$). For fair comparison we replaced only the feature projection part of the architecture from Figure~\ref{fig:global_architecture} with an LSS-based implementation, keeping the image feature extractor and task heads unchanged. 
Some inevitable adaptations have been applied to match input and output dimensions of the projection part during LSS implementation: extracted image features from the two deepest endpoints (4 and 5) have been fused via resizing; image feature channels has been adjusted for depth estimation. We observed a significant difference of 24.6 points in the F1-score between the two projection methods for a single-task \textit{Polygon-Yolo} experiment (our observed difference in performance using fisheye images is significantly greater compared to the reported difference when using pinhole images in \cite{zhou2022cvt}). The remaining rows of Table~\ref{table:Results_Ablation} demonstrates the cumulative benefits of applied methods on the top of our baseline ($\mathparagraph$): image top--cropping, multi-task learning (\textit{Polygon-Yolo} and \textit{segmentation}) and \gls{bev} data augmentations. Image-level data augmentations and roll rotations per camera did not contribute much, presumably because their potential contribution was already achieved by using pre-trained weights for backbones. 
It is also visible that flipping left-right ($\ddagger$) is the most effective augmentation in yaw axis, compared to mixing various rotation angles ($\dagger$).

\paragraph{Qualitative Results.}

Figure~\ref{fig:qualitative_results} illustrates our qualitative results. On the left of each subplot, \gls{svcs} cameras are shown. The middle illustration is the output of the \gls{mtfcvt}, where we see on a black background the following elements: the ego-vehicle is represented in the center (the center of the rear axis of the ego vehicle is the center of the illustration). Parking slot segmentation results are illustrated as heatmaps. Polygons are drawn around the detected parking slots. The green line of the polygon depicts it's entry line, while the rest of the lines are red. Its center point is distinguishable with a green dot. Corner dots are drawn in white if the network places them without a direct view, otherwise they are blue. the number tagged to a polygon is the computed confidence score. The vehicles are illustrated as filled red rectangles, where their heading is distinguishable with the green triangle zone. 
Finally, the video presented in the abstract shows the inference result of the \gls{mtfcvt} on an unseen vehicle \& camera rig, on a random parking area. This qualitative performance of the modal shows it's generalization capabilities to other camera rig than the ones used during the training process.

\section{Conclusion} \label{section:conclusion}

This work presents \gls{mtfcvt}, an innovative approach to parking area perception that accurately identifies both vacant parking spots and parked vehicles, with their correct orientation in a range of 25m × 25m around the ego-vehicle. \gls{mtfcvt} projects four fisheye \gls{svcs} into a \gls{bev} feature grid. It applies a cross-view multi-head attention mechanism to enhance overall scene understanding. Then, the multi-task learning of \textit{segmentation} and \textit{Yolo--Polygon}  detection is executed. Thanks to the used real 3D annotation, even our \textit{small} network configuration, positions objects with only 23 cm of error and achieves an f1-score of .86 in F1-score, outperforming both homograph stitching based method from \cite{wang2023holistic}, fisheye adaptation of LSS from \cite{philion2020lift} and the fisheye adaption of the initial CVT configuration from \cite{zhou2022cvt}. Running at a speed of 16 fps on an Nvidia Jetson Orin, \gls{mtfcvt} is suitable for low-speed parking applications. The proposed architecture demonstrates effective generalization capabilities to unseen vehicles and camera rigs successfully.

The limitation of \gls{mtfcvt} is its ignorance of \gls{bev} features of previous frames, in order to process temporal information and not being trained to recognize other parking relevant objects such as speed bumps or ground markings. 
Last, a new parking area dataset, including both the \gls{svcs} and the six-pinhole camera configuration should be collected, in order to compare detection performance of both configurations.

\section*{Acknowledgments}

The authors would like to especially thank Mihai Ilie, Shubham Chaudhari, Shubham Kajaria, Nitya Saxena and Manikandan Bakthavatchalam for creating and sharing the labels of the used dataset.




\bibliographystyle{apalike}

\bibliography{imvip}

\end{document}